\pdfoutput=1

\documentclass[11pt]{article}

\usepackage{acl}

\usepackage{times}
\usepackage{latexsym}

\usepackage[T1]{fontenc}

\usepackage[utf8]{inputenc}

\usepackage{microtype}

\usepackage{booktabs}
\usepackage{graphicx} 
\usepackage{graphics}
\usepackage{amssymb}
\usepackage{amsmath}
\usepackage{pifont}
\usepackage{makecell}
\usepackage{multirow}
\usepackage{colortbl}
\usepackage{enumitem}

\newcommand{\cmark}{\ding{51}}%
\newcommand{\xmark}{\ding{55}}%
\newcommand{\sectionref}[1]{\S\ref{#1}}
\usepackage[ruled,vlined]{algorithm2e}
\SetKwInput{KwInput}{Input}                
\SetKwInput{KwOutput}{Output}              

\title{On the Robustness of Reading Comprehension Models to \\Entity Renaming}

\author{Jun Yan$^{1}$ \quad Yang Xiao$^{2}$ \quad Sagnik Mukherjee$^{3}$ \quad Bill Yuchen Lin$^{1}$\\
\textbf{Robin Jia$^{1}$ \quad Xiang Ren$^{1}$}\\
University of Southern California$^{1}$ \quad Fudan University$^{2}$ \quad IIT Kanpur$^{3}$ \\
\texttt{\{yanjun,yuchen.lin,robinjia,xiangren\}@usc.edu}\\
\texttt{17307100059@fudan.edu.cn} \quad\texttt{sagnikm@iitk.ac.in}\\
}

\begin{document}
\maketitle

\begin{abstract}
We study the robustness of machine reading comprehension (MRC) models to entity renaming---do models make more wrong predictions when the same questions are asked about an entity whose name has been changed?
Such failures imply that models overly rely on entity information to answer questions, and thus may generalize poorly when facts about the world change or questions are asked about novel entities.
To systematically audit this issue, we present a pipeline to automatically generate test examples at scale, by replacing entity names in the original test sample with names from a variety of sources, ranging from names in the same test set, to common names in life, to arbitrary strings.
Across five datasets and three pretrained model architectures, MRC models consistently perform worse when entities are renamed, with particularly large accuracy drops on datasets constructed via distant supervision.
We also find large differences between models: SpanBERT, which is pretrained with span-level masking, is more robust than RoBERTa, despite having similar accuracy on unperturbed test data.
We further experiment with different masking strategies as the continual pretraining objective and find that entity-based masking can improve the robustness of MRC models.\footnote{Our code and data can be found at \url{https://github.com/INK-USC/entity-robustness}.}

\end{abstract}
\section{Introduction}
\label{sec:intro}

The task of machine reading comprehension (MRC) measures machines' understanding and reasoning abilities.
Recent research advances \citep{devlin2018bert, yang2019xlnet, khashabi2020unifiedqa} have driven MRC models to reach or even exceed human performance on several \textit{MRC benchmark datasets}.
However, their actual ability to solve the general \textit{MRC task} is still questionable \citep{kaushik2018much, sen2020models, sugawara2020assessing, lai2021machine}.
While humans show robust generalization on reading comprehension, existing works have revealed that MRC models generalize poorly to out-of-domain data distributions \citep{fisch2019mrqa} and are brittle under test-time perturbations \citep{pruthi2019combating, jia2019certified, jia2017adversarial}.
All these issues could naturally happen to MRC systems deployed in the wild, hindering them to make reliable predictions on user inputs with great flexibility.

\begin{figure}[t]
	\centering 
	\includegraphics[scale=0.44]{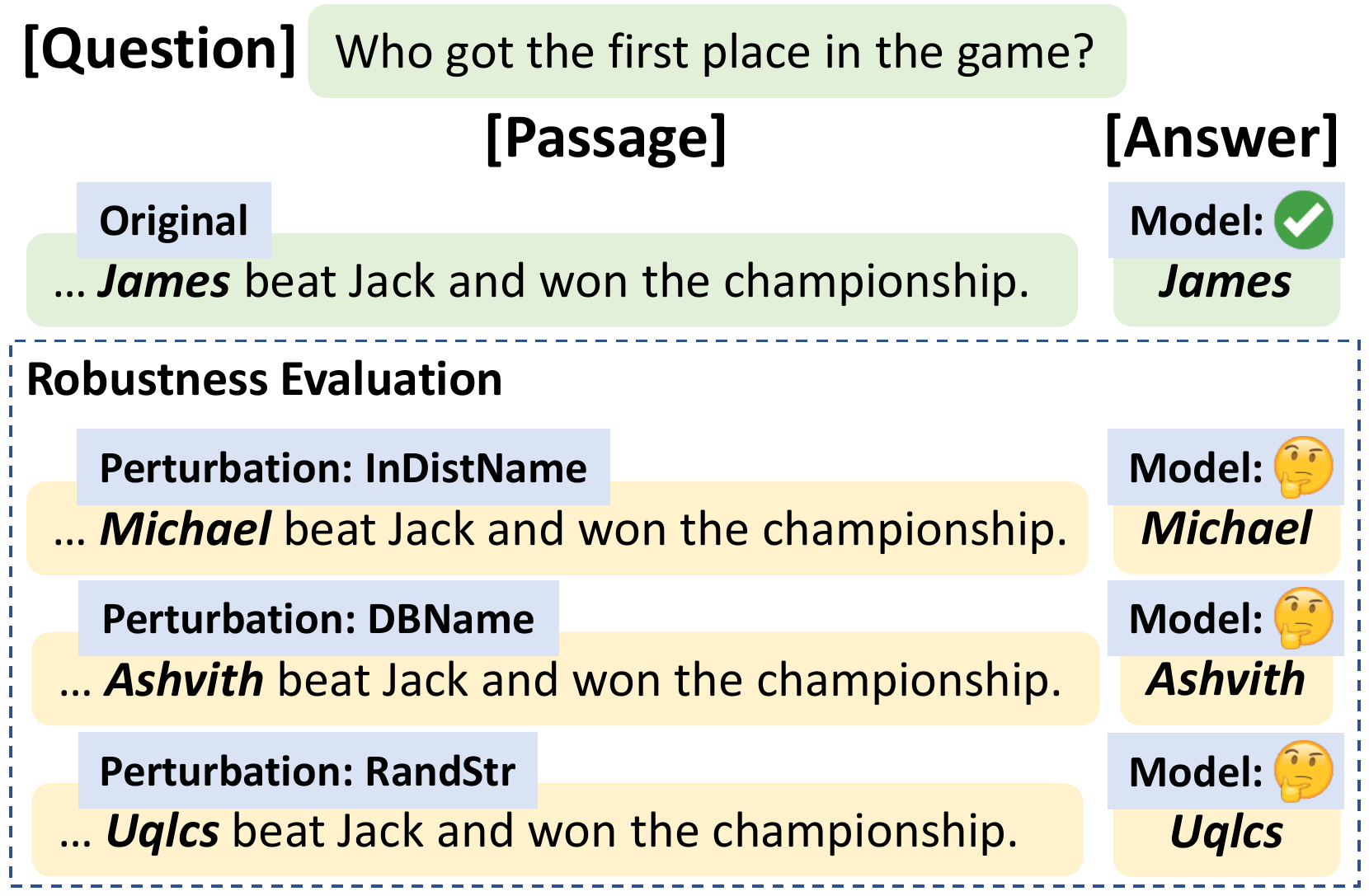}
	\caption{An illustrative example of the robustness to entity renaming and our proposed perturbations for robustness evaluation. ``Michael'' is from the answer of another test instance. ``Ashvith'' is a person name from an external database. ``Uqlcs'' is a random string with the same length as the original name.}
	\label{fig:illustration} 
\end{figure}

In this work, we focus on an important but understudied type of test-time distribution shift caused by novel entity (e.g., person and company) names.
Besides the evidence provided by the surrounding context, an MRC model also has the capacity to leverage the entity information to make predictions \citep{sugawara2018makes, chen2016thorough}.
The information associated with the entity name covers both world knowledge that can change over time and dataset shortcuts that are unlikely to generalize.
While contributing to performance on certain benchmarks, the over-reliance on specific entity names leads to an overestimation of model's actual ability to read and comprehend the provided passage \citep{penas2011overview}.
It also hinders model generalization to novel entity names, which itself is challenging due to the large space of valid entity names induced by the flexibility of entity naming. 
For example, person names can be chosen from a large vocabulary depending on the country, while companies can be named in an even more creative way, not to mention new names that are being invented every day.
As illustrated in Figure~\ref{fig:illustration}, keeping the reasoning context unchanged, a robust MRC model is supposed to correctly locate the same span of a named entity as the answer, even after it gets renamed.

To audit model robustness, we use entity renaming as test time perturbation to mimic the situation where a deployed MRC model encounters questions asking for novel entity names in the emerging data.
We design a general pipeline to generate natural perturbations of MRC instances by swapping the answer entity name with another valid name throughout the passage.
We design perturbation rules and collect resources for three types of entities with large name space: Person, Organization, and Geopolitical Entity.

With this proposed analysis framework, we conduct extensive experiments on five datasets and three pretrained language models.
Data-wise, we find that distantly supervised MRC datasets lead to less robustness.
Entity-wise, we find that geopolitical entities pose a greater challenge than people and organizations when renamed. 
Model-wise, we find that SpanBERT is more robust than BERT and RoBERTa, mainly due to its lower sensitivity to domain shift on names, which is likely a benefit of its span-focused pretraining objective.
Inspired by this, we investigate several continual pretraining objectives and find that an entity-based masking strategy can further improve robustness.
\section{Analysis Setup}
\label{sec:setup}

\subsection{Extractive MRC}
\label{subsec:task}

The task of MRC tests a machine' understanding and reasoning abilities by asking it to answer the question based on the provided passage.
We focus on extractive MRC, where the answer is a span in the passage.
Formally, given a \textbf{question} $Q$ and a \textbf{passage} $P$ of $n$ tokens $P=\{x_1 \ldots, x_n\}$, a model is expected to predict an \textbf{answer} span $a=\{x_i,\ldots,x_{i+k}\} (1\leq i\leq i+k\leq n)$ in the passage $P$ as a response to the question $Q$.
We use exact match (\textbf{EM}) as the metric for MRC evaluation, which is the percentage of test instances that the model exactly predicts one of the gold answers.

In both real-world scenarios and MRC datasets, a large portion of questions ask about entities like people, organizations and locations.
While unmentioned background knowledge about the entities might be helpful for solving the questions, overly relying on it makes the model hard to adapt to updated facts provided by the passage and generalize to novel entities.
Especially, we contrast MRC with closed-book QA, which requires a model to directly answer questions without access to any document passage.
Closed-book QA tests a model's ability to pack knowledge into its parameters and retrieve knowledge from parameters to answer the question.
On the contrary, we expect an MRC model to reason based on the provided passage.

\subsection{Evaluation Protocol}
\label{subsec:evaluation}

We study the robustness of MRC models via test-time perturbation.
Given an original test set $D_\text{test}$ and a perturbation function $f_\text{perturb}$ (detailed in \sectionref{sec:method}) as inputs, we construct $N$ perturbed test sets with $N$ \textit{perturbation seeds}.
We evaluate the model on the $N$ perturbed test sets.
By averaging the results, we get the \textbf{average-case EM score} as the final metric, which measures the average impact on the model performance caused by the names from a certain perturbation.
We set $N=5$ in experiments.

\subsection{Datasets}
\label{subsec:datasets}

We choose five datasets with different characteristics from the MRQA 2019 shared task \citep{fisch2019mrqa}: 
\textbf{SQuAD} \citep{rajpurkar2016squad}, \textbf{Natural Questions (NQ)} \citep{kwiatkowski2019natural}, \textbf{HotpotQA} \citep{yang2018hotpotqa}, \textbf{SearchQA} \citep{dunn2017searchqa}, and \textbf{TriviaQA} \citep{joshi2017triviaqa}.
Since the official test sets of the MRQA datasets are hidden, we use the development set as the in-house test set, and hold out 10\% of the training data as the in-house development set.
Their statistics are shown in Table~\ref{tab:datasets}.

\begin{table}[t]
\centering
\scalebox{0.8}
{
\begin{tabular}{@{}c|rrrc}
\toprule
Dataset & \# Train & \# Dev & \# Test & DS? \\
\midrule
SQuAD & 77,929 & 8,659 & 10,507 & \xmark \\
NQ & 84,577 & 9,367 & 12,836 & \xmark \\
HotpotQA & 65,636 & 7,292 & 5,901 & \xmark \\
SearchQA & 105,646 & 11,738 & 16,980 & \cmark \\
TriviaQA & 42,569 & 4,696 & 7,785 & \cmark \\
\bottomrule
\end{tabular}
}
\caption{Evaluation datasets. ``DS?'' indicates whether distant supervision is used for data collection.}
\label{tab:datasets}
\end{table}

As a major difference in data collection, SQuAD, NQ, and HotpotQA employ crowdworkers to annotate the answer span in the passage, while SearchQA and TriviaQA use distant supervision to match the passage with the question.
Distant supervision provides no guarantee that the passage contains enough evidence to derive the answer.
The context where the entity span shows up may not even be related to the question.

\subsection{MRC Models}
\label{subsec:models}

We experiment with three pretrained language models that have demonstrated strong performance on popular MRC benchmarks.
\textbf{BERT} \citep{devlin2018bert} is trained on English Wikipedia plus BookCorpus with masked language modeling (MLM) and next sentence prediction (NSP) as self-supervised objectives.
\textbf{RoBERTa} \citep{liu2019roberta} improves over BERT mainly by dropping the NSP objective and increasing the pretraining time and the size of pretraining data.
\textbf{SpanBERT} \citep{joshi2020spanbert} masks random contiguous spans to implement MLM and replaces NSP with a span-boundary objective (SBO).

The pretrained language models are finetuned on the MRC dataset to predict the start and end tokens of the answer span based on the concatenated question and passage \citep{devlin2018bert}.
By default, all pretrained language models in the main experiments are case-sensitive and in their \texttt{BASE} sizes.
More details are shown in Appendix~\sectionref{app:training}.

\section{Entity Name Substitution}
\label{sec:method}

\begin{figure*}[t]
	\centering 
	\includegraphics[scale=0.41]{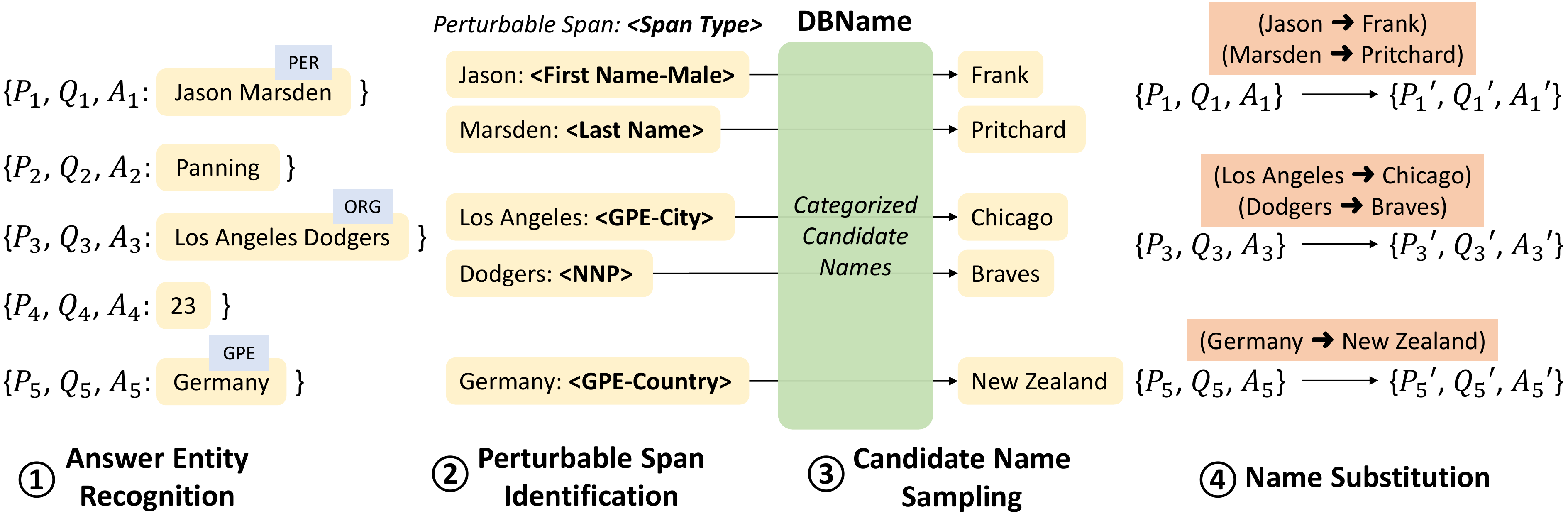}
	\caption{The perturbation pipeline for performing entity name substitution on MRC instances.}
	\label{fig:pipeline} 
\end{figure*}

In this section, we introduce our method for perturbing an MRC test set with substitution entity names, i.e., the instantiation of $f_\text{perturb}$.
Generating substitution names is at the core of our evaluation as different kinds of names measure a model's behavior in different situations with different robustness implications.
We propose three categories of perturbations on three entity types and collect the corresponding name resources, aiming to audit a model's robustness from different perspectives.

\subsection{Perturbation Pipeline}
\label{subsec:pipeline}

As illustrated in Figure~\ref{fig:pipeline}, our perturbation pipeline consists of four steps, which are introduced below.

\paragraph{Step 1: Answer Entity Recognition.}
As we focus on the effect of answer entity renaming, we first identify entities in the answers by performing named entity recognition (NER) with spaCy \citep{spacy} on the passage and extract the results on the answer spans.
We identify three types of named entities: Person (\textbf{PER}), Organization (\textbf{ORG}), and Geopolitical Entity (\textbf{GPE}).
All of them frequently appear as answers and have large space of valid names, making it important and challenging for models to robustly handle.

\paragraph{Step 2: Perturbable Span Identification.}
To facilitate name substitution, we assign metadata to detected entity names by identifying \textit{perturbable spans} within the entity name. 
For each type of entity names, we define the applicable \textit{span types} in Table~\ref{tab:matadata}.
The heuristics for identifying each type of perturbable spans are introduced in Appendix~\sectionref{app:perturbable}. 
Note that given one or more entity types of interest, in this step we filter the test data to only keep a subset of instances with non-empty metadata for the corresponding entity types, which are instances that are ready to be perturbed. 
Sizes of the perturbable subsets for different entity types and their union (\textbf{MIX}) are shown in Table~\ref{tab:perturbable}.

\begin{table}[t]
\centering
\scalebox{0.7}
{
\begin{tabular}{@{}ccl@{}}
\toprule
\multicolumn{2}{c|}{Entity Type}                                                       & \multicolumn{1}{c}{Applicable Types of Perturbable Spans}  \\ \midrule
\multicolumn{2}{c|}{\multirow{4}{*}{\makecell{PER\\(4)}}}                                             & <First Name-Male> (e.g., Richard, Morton)     \\
\multicolumn{2}{c|}{}                                                                 & <First Name-Female> (e.g., Lauren, Jennifer)   \\
\multicolumn{2}{c|}{}                                                                 & <First Name-Neutral> (e.g., Shine, Frankie) \\
\multicolumn{2}{c|}{}                                                                 & <Last Name> (e.g., Marx, Winfrey)          \\ 
\midrule
\multicolumn{1}{c|}{\multirow{5}{*}{\makecell{ORG\\(5)}}} & \multicolumn{1}{l|}{\multirow{2}{*}{}}    & <NNP> (e.g., Celtic, Tiffany)                \\
\multicolumn{1}{c|}{}                     & \multicolumn{1}{l|}{}                     & <Rare> (e.g., Hufflepuff, Pokemon)               \\ \cmidrule(lr){2-3}
\multicolumn{1}{c|}{}                     & \multicolumn{1}{c|}{\multirow{3}{*}{\makecell{GPE\\(3)}}} & <GPE-Country> (e.g., Iceland, Algeria)        \\
\multicolumn{1}{c|}{}                     & \multicolumn{1}{c|}{}                     & <GPE-State> (e.g., New Brunswick, Ohio)          \\
\multicolumn{1}{c|}{}                     & \multicolumn{1}{c|}{}                     & <GPE-City> (e.g., Boston, Sonsonate)           \\ \bottomrule
\end{tabular}
}
\caption{Applicable metadata for each entity type in the perturbation pipeline.}
\label{tab:matadata}
\end{table}

\begin{table}[t]
\centering
\scalebox{0.8}
{
\begin{tabular}{@{}c|rrr|r}
\toprule
Dataset & \# PER & \# ORG & \# GPE & \# MIX \\
\midrule
SQuAD & 1,170 & 1,095 & 602 & 2,613 \\
NQ & 3,257 & 1,207 & 1,414 & 5,150 \\
HotpotQA & 1,351 & 824 & 788 & 2,614 \\
SearchQA & 5,707 & 2,450 & 2,248 & 8,688 \\
TriviaQA & 2,747 & 1,276 & 1,270 & 4,351 \\
\bottomrule
\end{tabular}
}
\caption{Statistics of the perturbable subsets for different entity types and their union (``MIX'').}
\label{tab:perturbable}
\end{table}

\paragraph{Step 3: Candidate Name Sampling.}

For each perturbable span, we get its substitution name by querying an external dictionary with the span type.
The substitution name is randomly sampled from a pool of names in the external dictionary with the same span type.
We collect dictionaries with names of different characteristics serving for different analysis purpose, which are detailed in \sectionref{subsec:candidate_names}.

\paragraph{Step 4: Name Substitution.}

Once we have a candidate name for each perturbable span, we perform string mapping on the passage, question, and the gold answer, to finish the entity renaming in MRC instances.
The name substitution changes \textit{all} mentions of the answer entity in the passage while keeping the other reasoning context.

\subsection{Candidate Name Collection}
\label{subsec:candidate_names}

We consider three types of candidate names for perturbations in our main experiments to simulate the domain shift of entity names during test time.

\paragraph{In-Distribution Name (InDistName).}

The set of candidate names with their span types is the same as the perturbable spans with their types identified from the gold answers in the test set.
This ensures that no new name is introduced to the test set.

\paragraph{Database Name (DBName).}

We collect names in the real world by referring to relevant databases. For PER, we collect first names\footnote{\scriptsize \url{https://www.ssa.gov/oact/babynames/limits.html}} (with gender frequency) and last names\footnote{\scriptsize\url{https://www.census.gov/topics/population/genealogy/data/2010\_surnames.html}} from the official statistics of person names in the U.S. (We experiment with names from other countries and languages in \sectionref{subsec:difficulty}.)
We regard a first name as a male/female name if its male/female frequency is two times larger than its frequency of the opposite gender.
The remaining names are considered as neutral.
Following the practice for identifying perturbable spans, we get the list of country/state/city names using \textit{Countries States Cities Database} and the NNP list using PTB.
Rare words constitute an open vocabulary so they will not be substituted under the DBName perturbation.

\paragraph{Random String (RandStr).}

The RandStr perturbation is different from the other two as it neglects the query span type when preparing the candidates.
We generate a random alphabetical string of the same length and casing as the original perturbable span.
Names from low-resource languages can look quite irregular to the pretrained language models.
Random string as an extreme case provides an estimation of the performance in this scenario.

\subsection{Perturbation Quality}

\begin{table}[t]
\centering
\scalebox{0.8}
{
\begin{tabular}{@{}c|ccc}
\toprule
Accuracy & PER & ORG & GPE \\
\midrule
Perturbable Span Identification & 93.3\% & 86.7\% & 93.3\% \\
Name Substitution & 86.7\% & 96.7\% & 96.7\% \\
\bottomrule
\end{tabular}
}
\caption{Validity of the two key steps in the perturbation pipeline on 30 randomly sampled TriviaQA instances for each entity type.}
\label{tab:validity}
\end{table}

The validity of the perturbed instances depends on the quality of the perturbation pipeline (\sectionref{subsec:pipeline}).
We manually check the accuracy of the perturbation steps on TriviaQA, which demonstrates the largest performance drop as we will show.
Out of the four steps in the pipeline (Figure~\ref{fig:pipeline}), we evaluate the accuracy of step 2 (``Perturbable Span Identification'') and step 4 (``Name Substituion'') while the accuracy of the other two steps can be inferred.
The evaluation details are provided in Appendix~\sectionref{app:eval_perturbation}.
As shown in Table~\ref{tab:validity}, our method gets acceptable accuracy on the three entity types, confirming the quality of the perturbed test sets.

\section{Results and Analysis}
\label{sec:results}

\begin{figure*}[t]
	\centering 
	\includegraphics[scale=0.9]{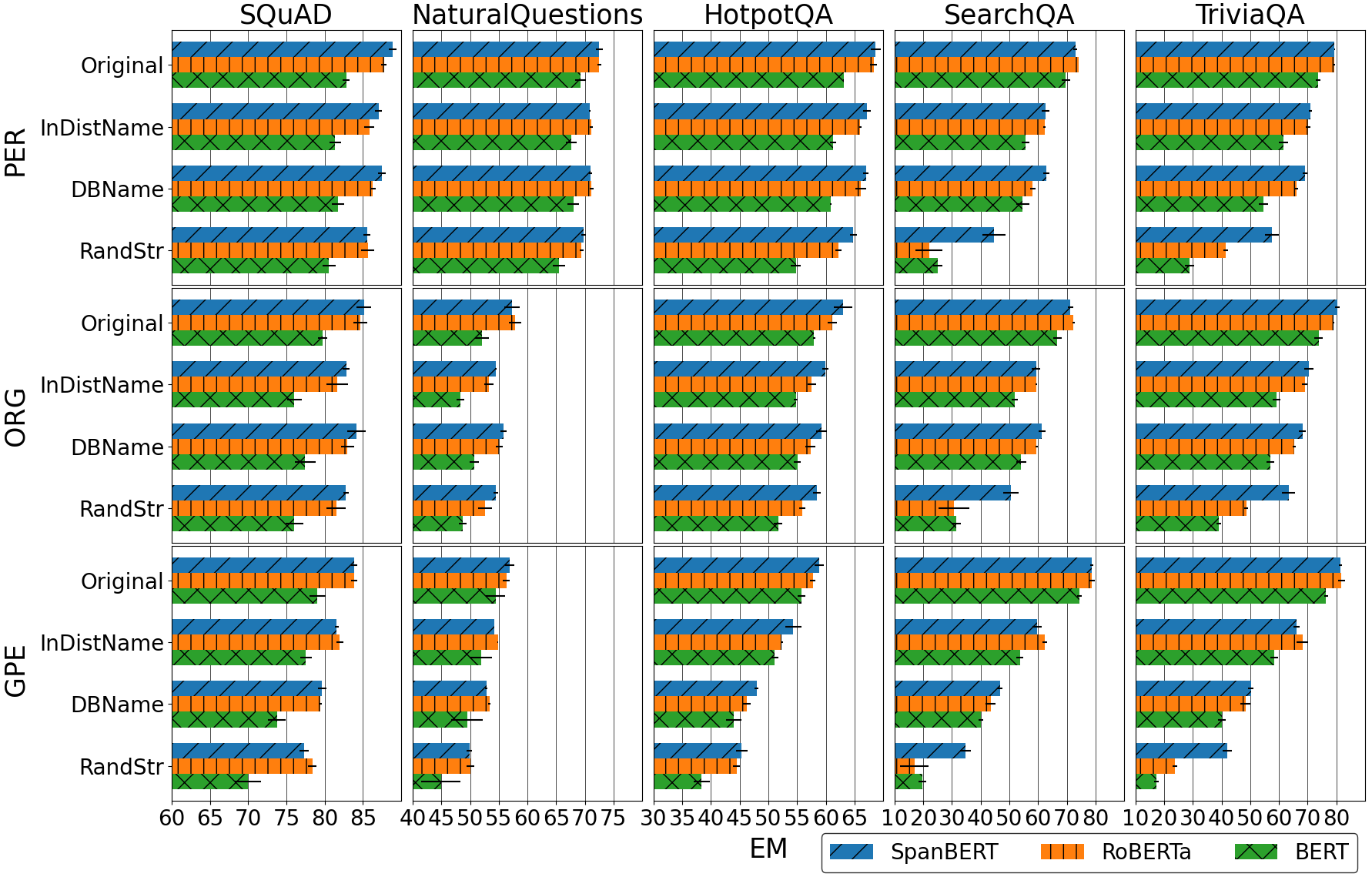}
	\caption{\textbf{Main results.} EM scores for MRC models evaluated on datasets under different perturbations.}
	\label{fig:main_table} 
\end{figure*}

The average-case EM scores on the original and perturbed test sets are presented in Figure~\ref{fig:main_table}.
We report the mean and standard deviation over 3 \textit{training seeds}.
We analyze the results from several angles by aggregating across certain dimensions.

\begin{table}[t]
\centering
\scalebox{0.63}
{
\begin{tabular}{@{}c|ccc|cc}
\toprule
BERT@MIX & SQuAD & NQ & HotpotQA & SearchQA & TriviaQA \\
\midrule
Original & $81.2_{\pm 0.3}$ & $64.4_{\pm 1.0}$ & $60.0_{\pm 0.2}$ & $69.5_{\pm 1.1}$ & $73.4_{\pm 0.8}$ \\
\midrule
InDistName & $78.7_{\pm 0.6}$ & $62.0_{\pm 1.2}$ & $\mathbf{56.8}_{\pm 0.4}$ & $\mathbf{53.6}_{\pm 1.3}$ & $\mathbf{59.0}_{\pm 1.4}$ \\
DBName & $\mathbf{78.8}_{\pm 0.9}$ & $\mathbf{62.1}_{\pm 1.3}$ & $54.9_{\pm 0.3}$ & $50.2_{\pm 1.8}$ & $50.4_{\pm 1.6}$ \\
RandStr & $76.9_{\pm 1.0}$ & $59.0_{\pm 1.7}$ & $49.5_{\pm 0.8}$ & $23.6_{\pm 1.2}$ & $25.4_{\pm 1.4}$ \\

\bottomrule
\end{tabular}
}
\caption{\textbf{Comparison of different datasets.} EM scores of BERT on the original and perturbed test sets of the MIX entity type.}
\label{tab:compare_datasets}
\end{table}

\begin{table}[t]
\centering
\scalebox{0.63}
{
\begin{tabular}{@{}c|ccc|cc}
\toprule
\makecell{Wrong Entity\\Error} & SQuAD & NQ & HotpotQA & SearchQA & TriviaQA \\
\midrule
Original & 33.9\% & 34.3\% & 27.3\% & 46.3\% & 69.0\% \\
\midrule
InDistName & 38.4\% & 37.5\% & 32.3\% & 66.2\% & 76.6\% \\
DBName & 38.0\% & 37.3\% & 33.3\% & 67.2\% & 77.6\% \\
RandStr & 42.7\% & 41.2\% & 39.1\% & 84.7\% & 86.8\% \\

\bottomrule
\end{tabular}
}
\caption{\textbf{Error analysis.} The percentage of wrong entity errors of BERT on the original and perturbed test sets of the MIX entity type.}
\label{tab:error}
\end{table}

\subsection{Which datasets lead to less robustness?}

\textbf{Training on MRC datasets created with distant supervision leads to less robustness.}
In Table \ref{tab:compare_datasets}, we show the results of BERT on the original and perturbed test sets, while results of RoBERTa and SpanBERT follow similar patterns. 
The perturbations on all 3 entity types are combined (shown as ``MIX'').
We find that models trained on SQuAD, NQ, and HotpotQA (with at most 6\% performance drop under the DBName perturbation) are significantly more robust than models trained on SearchQA and TriviaQA (with about 20\% performance drop under the DBName perturbation).
While the first group of datasets are human-labeled, the later group of datasets are constructed using distant supervision.
Such correlation indicates that training noise due to mismatched questions and passages harms model's robustness.
We hypothesize the reason to be that, the passage in the human-annotated datasets usually provides enough evidence to derive the answer, so a model is able to learn the actual task of ``reading comprehension'' from the data.
On the contrary, SearchQA and TriviaQA use web snippets as the source of passages.
The labeling process of distant supervision assumes that ``the presence of the answer string implies the document \textit{does} answer the question'' \citep{joshi2017triviaqa}, while the document may or may not contain all facts needed to support the answer.
In this case, because the actual reading comprehension task is difficult to learn due to lack of evidence, the model could be prone to use entity-specific background knowledge (e.g. assuming that ``Jack Higgins'' is a British author regardless of the context) or learn dataset-specific shortcuts associated with certain names via memorization (e.g., choosing ``Jack Higgins'' whenever it's mentioned in the passage and the question asks for an author), which causes the robustness issue.

To better understand the failure cases, we categorize the errors made by the model into two classes: wrong entity errors and wrong boundary errors, based on whether the predicted span has any word overlap with the gold answer.
We report the percentage of wrong entity errors in Table~\ref{tab:error}.
On all datasets, wrong entity errors make up a larger percentage of all errors when the test sets get perturbed.
This suggests that the performance drop is mainly caused by the increasing errors in identifying the correct answer entity rather than accurately predicting the boundary of a correctly-identified answer entity.

\begin{table}[t]
\centering
\scalebox{0.63}
{
\begin{tabular}{@{}c|ccc|cc}
\toprule
BERT & SQuAD & NQ & HotpotQA & SearchQA & TriviaQA \\
\midrule
PER-Original & $82.8_{\pm 0.4}$ & $69.3_{\pm 0.9}$ & $63.1_{\pm 0.1}$ & $69.7_{\pm 1.4}$ & $73.6_{\pm 0.7}$ \\
PER-DBName & $81.7_{\pm 0.8}$ & $68.0_{\pm 1.0}$ & $60.8_{\pm 0.2}$ & $54.6_{\pm 2.3}$ & $54.6_{\pm 1.6}$ \\
PER-$\Delta$ & 1.1 & 1.3 & 2.3 & 15.1 & 19.0 \\
\midrule
ORG-Original & $79.7_{\pm 0.6}$ & $52.1_{\pm 1.2}$ & $58.0_{\pm 0.3}$ & $66.7_{\pm 1.5}$ & $73.8_{\pm 1.5}$ \\
ORG-DBName & $77.5_{\pm 1.4}$ & $50.8_{\pm 0.8}$ & $55.0_{\pm 0.6}$ & $54.2_{\pm 1.6}$ & $57.0_{\pm 1.5}$ \\
ORG-$\Delta$ & 2.2 & 1.3 & 3.0 & 12.5 & 16.8 \\
\midrule
GPE-Original & $79.1_{\pm 1.0}$ & $54.5_{\pm 1.7}$ & $55.8_{\pm 0.6}$ & $74.4_{\pm 0.8}$ & $76.4_{\pm 0.6}$ \\
GPE-DBName & $73.7_{\pm 1.1}$ & $49.5_{\pm 2.7}$ & $43.9_{\pm 1.3}$ & $40.1_{\pm 0.8}$ & $40.1_{\pm 1.3}$ \\
GPE-$\Delta$ & \textbf{5.4} & \textbf{5.0} & \textbf{11.9} & \textbf{34.3} & \textbf{36.3} \\
\bottomrule
\end{tabular}
}
\caption{\textbf{Comparison of different entity types.} EM scores of BERT on the Original and DBName test sets.}
\label{tab:compare_types}
\end{table}

\subsection{Which entity types are more challenging?}
\label{subsec:entity_type}

\textbf{GPE renaming poses the greatest robustness challenge. The renaming of PER and ORG are similarly less challenging.}
In Table~\ref{tab:compare_types}, we present the performance drop caused by the DBName perturbation for each entity type.
GPE renaming shows the largest performance drop.
The comparison of PER and ORG differs across datasets, but their corresponding performance drops are generally similar.
The reason is likely to be that the model is only exposed to a small number of distinct GPE names during finetuning compared to PER and ORG.
In the training set of TriviaQA, there are 40k ORG names and 54k PER names, but only 12k GPE names.
The lack of seen names makes it hard to learn the generalization ability.

\begin{table}[t]
\centering
\scalebox{0.65}
{
\begin{tabular}{@{}c|ccc}
\toprule
MIX & BERT & RoBERTa & SpanBERT \\
\midrule
Original & $69.5_{\pm 1.1}$/$73.4_{\pm 0.8}$ & $\mathbf{74.1}_{\pm 0.2}$/$78.6_{\pm 0.4}$ & $73.2_{\pm 0.7}$/$\mathbf{79.1}_{\pm 0.1}$\\
\midrule
InDistName & $53.6_{\pm 1.3}$/$59.0_{\pm 1.4}$ & $\mathbf{60.7}_{\pm 0.4}$/$67.8_{\pm 1.1}$ & $60.3_{\pm 1.4}$/$\mathbf{68.3}_{\pm 0.8}$ \\
DBName & $50.2_{\pm 1.8}$/$50.4_{\pm 1.6}$ & $54.0_{\pm 1.0}$/$60.5_{\pm 0.9}$ & $\mathbf{57.9}_{\pm 1.0}$/$\mathbf{63.1}_{\pm 0.8}$ \\
RandStr & $23.6_{\pm 1.2}$/$25.4_{\pm 1.4}$ & $21.0_{\pm 4.8}$/$35.6_{\pm 0.2}$ & $\mathbf{41.5}_{\pm 3.2}$/$\mathbf{51.9}_{\pm 2.3}$\\

\bottomrule
\end{tabular}
}
\caption{\textbf{Comparison of different models.} EM scores on the original and perturbed test sets of the MIX entity type on SearchQA/TriviaQA.}
\label{tab:compare_models}
\end{table}

\subsection{Which models are more robust?}
\textbf{On distantly supervised datasets, SpanBERT is more robust than RoBERTa, which is more robust than BERT.}
In Table~\ref{tab:compare_models}, we show the performance of the three models under perturbations of the MIX entity type on SearchQA and TriviaQA.
While RoBERTa and SpanBERT show comparable performance on the original and InDistName test sets, SpanBERT's improvement over RoBERTa becomes larger with more difficult perturbations.
Meanwhile, BERT shows even larger performance decreases than RoBERTa.
The models' performance differences are mainly attributed to their different pretraining strategies.
RoBERTa's improvement over BERT indicates that a better pretraining configuration (as measured by the performance on the in-domain original test set) is also beneficial to the performance on the perturbed test sets, suggesting better generalization ability to the out-of-domain data.
This correlation is consistent with the findings in \citet{miller2021accuracy}.
SpanBERT's particular advantage on the perturbed test sets indicates its span-focused pretraining objective (span-based MLM and span prediction based on boundary tokens) is especially helpful for the span-related robustness, which is desired for the MRC task.

\begin{table}[t]
\centering
\scalebox{0.65}
{
\begin{tabular}{@{}c|ccc}
\toprule
MIX & BERT & RoBERTa & SpanBERT \\
\midrule
Original & $73.4_{\pm 0.8}$/$76.3_{\pm 1.1}$ & $78.6_{\pm 0.4}$/$82.3_{\pm 0.2}$ & $\mathbf{79.1}_{\pm 0.1}$/$\mathbf{82.8}_{\pm 0.6}$\\
\midrule
InDistName & $59.0_{\pm 1.4}$/$61.3_{\pm 0.4}$ & $67.8_{\pm 1.1}$/$70.8_{\pm 0.6}$ & $\mathbf{68.3}_{\pm 0.8}$/$\mathbf{72.1}_{\pm 0.7}$ \\
DBName & $50.4_{\pm 1.6}$/$52.6_{\pm 0.7}$ & $60.5_{\pm 0.9}$/$62.6_{\pm 1.3}$ & $\mathbf{63.1}_{\pm 0.8}$/$\mathbf{66.9}_{\pm 0.6}$ \\
RandStr & $25.4_{\pm 1.4}$/$27.9_{\pm 2.4}$ & $35.6_{\pm 0.2}$/$35.8_{\pm 3.1}$ & $\mathbf{51.9}_{\pm 2.3}$/$\mathbf{53.6}_{\pm 4.9}$\\
\bottomrule
\end{tabular}
}
\caption{\textbf{Comparison of different model sizes.} EM scores of the \texttt{BASE}/\texttt{LARGE} variants of models on the original and perturbed test sets of the MIX entity type on TriviaQA.}
\label{tab:compare_sizes}
\end{table}

\textbf{Larger models are not more robust.}
In Table~\ref{tab:compare_sizes}, we compare the performance of the \texttt{BASE} and \texttt{LARGE} variants of pretrained models on TriviaQA.
The performance drops from the original test sets to the perturbed test sets are similar for these two variants in most cases, suggesting that simply increasing the model size can not resolve the robustness issue.

\subsection{How can we disentangle reasons for performance drop?}
\label{subsec:disentangle}
\textbf{Both loss of entity knowledge and domain shift on names happen during renaming.}
The information associated with the entity name that can be leveraged by the model includes both entity knowledge and name clues.
\textbf{Entity knowledge} refers to the \textit{world knowledge} associated with \textit{the referred entity}, like ``Michelle Obama is the wife of Barack Obama,'' while \textbf{name clues} refer to \textit{statistical clues} associated with the name's \textit{surface form}, like ``Barack Obama is likely to be a male name'', ``Barack Obama as an in-distribution name is likely to be the answer for this dataset''.
While all perturbations break the entity knowledge about the original entity, InDistName doesn't introduce additional domain shift on names and largely preserve the name clues.
Going from InDistName to other perturbations, the substitution names become more and more out of the dataset distribution. 
This performance drop can be attributed to the model's sensitivity to name-related domain shift.

\begin{table}[t]
\centering
\scalebox{0.8}
{
\begin{tabular}{@{}c|ccc}
\toprule
Unseen@MIX & \makecell{Original/\\InDistName} & DBName & RandStr \\
\midrule
SQuAD & 33\% / 24\% & 94\% / 92\% & 100\% / 100\% \\
NQ & 14\% / 4\% & 95\% / 89\% & 100\% / 100\% \\
HotpotQA & 19\% / 3\% & 96\% / 89\% & 100\% / 100\% \\
SearchQA & 7\% / 0\% & 92\% / 78\% & 100\% / 100\% \\
TriviaQA & 21\% / 2\% & 93\% / 83\% & 100\% / 100\% \\
\bottomrule
\end{tabular}
}
\caption{The percentage of test answer entity tokens that are never seen in the training answer/passage entities.}
\label{tab:domain_shift_token}
\end{table}

\begin{table}[t]
\centering
\scalebox{0.8}
{
\begin{tabular}{@{}c|cccc}
\toprule
Accuracy & Original & InDistName & DBName & RandStr \\
\midrule
PER & 75.6\% & 73.4\% & 58.0\% & 60.0\% \\
ORG & 60.1\% & 50.6\% & 49.2\% & 48.0\% \\
GPE & 84.4\% & 83.4\% & 43.2\% & 27.6\% \\
\bottomrule
\end{tabular}
}
\caption{Accuracy of the trained NER model to recognize the original and perturbed answer entities on SQuAD.}
\label{tab:domain_shift_entity}
\end{table}

We adopt two measurements to better understand the domain shift on names.
As a token-level measurement, we calculate the percentage of test answer entity tokens that are never seen in entities in training answers and entities in training passages, as shown in Table~\ref{tab:domain_shift_token}.
Different datasets have different percentages of unseen tokens in the original test sets, which are mainly affected by the size and diversity of training data.
The number goes up with the DBName and RandStr perturbations.
As an entity-level measurement, we train an NER model\footnote{\url{https://spacy.io/usage/training}} on the training passages, with named entities annotated by spaCy as ground truth. We evaluate the trained model on perturbed test sets and calculate its accuracy of recognizing the perturbed answer entity.
The results are shown in Table~\ref{tab:domain_shift_entity}.
As a sign of domain shift, the recognition of answer entities become more difficult when they get perturbed.
GPE shows the most significant perturbation drop, which correlate with our observation on the MRC task (\sectionref{subsec:entity_type}).

\textbf{SpanBERT's superior robustness over RoBERTa is mainly from handling domain shift.}
From SearchQA and TriviaQA results in Table~\ref{tab:compare_models}, we find that RoBERTa and SpanBERT rely similarly on the entity knowledge (\textasciitilde 13\% performance drop from Original to InDistName on SearchQA and \textasciitilde 11\% on TriviaQA).
SpanBERT's advantage over RoBERTa is mainly on its good robustness to domain shift on names, shown by the perfromance drop from IndistName to other perturbations.
BERT relies slightly more on entity knowledge but much more sensitive to domain shift on names.

\subsection{Bias Exhibited by Person Names}
\label{subsec:difficulty}

\paragraph{National Origins.}
As the DBName perturbation uses person names in the U.S., it cannot fully reflect the model's robustness behavior when encountering real-world names of different national origins.
Therefore, we additional collect names from more countries (India, China) and languages (French, Arabic) to study the potential bias in MRC models.
We use the romanized form of names.
Table~\ref{tab:cultural} shows the performance comparison of models when evaluated with the person names from different countries and languages on SearchQA and TriviaQA.
Names form the U.S. and French-speaking countries generally achieve the highest EM scores.
Names from China get the lowest performance for the most of time, with significant EM drop (8.4\% on SearchQA and 9.8\% on TriviaQA for BERT) from U.S. names.
The performance gap between different countries and languages becomes smaller with more robust models.

\paragraph{Other Factors.}
We also consider other factors of a name that could be related to biased model performance.
We limit our scope to the U.S. first names and sample 1500 names from the database.
We consider two features for each name.
\textit{Gender polarity} is defined as $\texttt{max}(\frac{f_m}{f_f},\frac{f_f}{f_m})$, where $f_m$, $f_f$ are the male frequency and female frequency of a name provided by the database.
It measures the gender ambiguity of the name. 
\textit{Popularity} is defined as $f_m+f_f$.
We calculate the EM score for a name by evaluating on a test set where all answer first names get replaced with this name.
For what we have tried, we didn't find evidence to support a correlation between each factor and the EM score.
For example, with SpanBERT on TriviaQA, names with top 20\% gender polarity gets 72.7\% EM on average; while the bottom 10\% names gets 72.8\% EM.
The numbers are 73.0\% vs 72.7\% for popularity.
We leave exploring factors that correlate with the difficulty of a name as future work.

\begin{table}[t]
\centering
\scalebox{0.65}
{
\begin{tabular}{@{}c|ccc}
\toprule
\makecell{Country/\\Language} & BERT & RoBERTa & SpanBERT \\
\midrule
U.S. & $54.6_{\pm 2.3}$/$54.6_{\pm 1.6}$ & $58.1_{\pm 0.9}$/$\mathbf{66.1}_{\pm 0.6}$ & $63.0_{\pm 1.1}$/$\mathbf{69.1}_{\pm 0.7}$ \\
French & $\mathbf{55.5}_{\pm 2.2}$/$\mathbf{56.1}_{\pm 1.7}$ & $\mathbf{58.2}_{\pm 1.1}$/$66.0_{\pm 0.5}$ & $63.0_{\pm 1.2}$/$68.8_{\pm 0.9}$ \\
India & $53.5_{\pm 2.5}$/$51.9_{\pm 2.7}$ & $56.5_{\pm 1.9}$/$63.9_{\pm 0.8}$ & $\mathbf{63.0}_{\pm 1.1}$/$68.0_{\pm 0.4}$ \\
Arabic & $53.3_{\pm 3.1}$/$48.8_{\pm 3.2}$ & $56.3_{\pm 2.1}$/$61.8_{\pm 0.9}$ & $62.8_{\pm 1.0}$/$66.2_{\pm 0.8}$ \\
China & $46.2_{\pm 2.5}$/$44.8_{\pm 3.6}$ & $54.0_{\pm 0.8}$/$63.0_{\pm 1.4}$ & $59.3_{\pm 2.0}$/$65.2_{\pm 0.4}$ \\
\midrule
RandStr & $25.0_{\pm 1.6}$/$28.9_{\pm 1.6}$ & $22.0_{\pm 4.7}$/$41.3_{\pm 0.8}$ & $44.6_{\pm 4.0}$/$57.4_{\pm 2.4}$ \\
\bottomrule
\end{tabular}
}
\caption{\textbf{Performance comparison of person names of different national origins.} EM scores on the original and perturbed test sets of the PER entity type on SearchQA/TriviaQA.}
\label{tab:cultural}
\end{table}

\subsection{Improving Robustness with Continual Pretraining}
SpanBERT's advantage over BERT suggests that some variants of MLM could be helpful for model robustness.
To further improve the robustness of SpanBERT, we adopt a training paradigm with an inserted continual pretraining stage and compare MLM with different masking strategies as the objectives.

\paragraph{Training Paradigm.}
\label{sec:defense:paradigm}
Existing works mainly seek to improve model robustness during finetuning with strategies like data augmentation \citep{ribeiro2019red, min2020syntactic}, but they usually increase finetuning time and requires additional data.
Some recent works \citep{gururangan2020don, ye2021influence} have explored improving a pretrained language model with ``continual pretraining''---continuing to train a pretrained model for more steps with some objective.
The generated checkpoint can be used for finetuning on any dataset in the \textit{standard} way with no additional cost.

\paragraph{Experimental Setup.}
\label{sec:defense:setup}
The masking policy in MLM plays an important role in instructing model learning, which can be potentially used to improve model robustness.
Inspired by previous works, we experiment with four heuristic masking policies to implement the MLM objective: 
\textbf{MLM (vanilla)}, \textbf{MLM (whole word)}, \textbf{MLM (span)}, and \textbf{MLM~(entity)}.
They perform masking at token, whole-word, span, and entity level respectively.
Starting from SpanBERT (-\texttt{BASE}), we run continual pretraining with the above objectives for 8,000 steps.
More details are described in Appendix~\sectionref{app:continual}.

\paragraph{Results.}
\label{sec:defense:results}

The results for models finetuned from SpanBERT and different continually pretrained models are shown in Table~\ref{tab:results_defense}.
On SQuAD, all masking policies slightly downgrade the performance.
With not much room for robustness improvement, running continual pretraining is probably at the cost of slightly sacrificing the performance due to the inconsistent objective and discontinuous learning rate that are applied when starting the continual pretraining.
On SearchQA and TriviaQA, out of the four masking policies, the entity-based masking policy shows consistent improvement over SpanBERT.
As analyzed in \sectionref{subsec:disentangle}, name-related domain shift is a major challenge for the model to handle.
By predicting the masked entity, the model is exposed to the diverse entities in the pertraining corpus in a more explicit way, and gain a better sense in recognizing entities.
Note that the improvement is not statistically significant in some cases and we leave the exploration of more effective methods to improve model robustness as future work.

\begin{table*}[t]
\centering
\scalebox{0.71}
{
\begin{tabular}{@{}l|ccc|ccc|ccc@{}}
\toprule
         & \multicolumn{3}{c|}{SQuAD}                & \multicolumn{3}{c|}{SearchQA} & \multicolumn{3}{c}{TriviaQA}              \\
         \cmidrule(l){2-10}
         Model / Perturbation (MIX) & Original & DBName & RandStr & Original & DBName & RandStr & Original & DBName & RandStr \\
         \midrule
SpanBERT & $\mathbf{86.8}_{\pm 0.5}$ & $\mathbf{84.9}_{\pm 0.4}$ & $\mathbf{83.0}_{\pm 0.1}$ & $73.2_{\pm 0.7}$ & $57.9_{\pm 1.0}$ & $41.5_{\pm 3.2}$ & $79.1_{\pm 0.1}$ & $63.1_{\pm 0.8}$ & $51.9_{\pm 2.3}$ \\
\midrule
SpanBERT w/ continual pretraining  & \cellcolor{gray!50} & \cellcolor{gray!50} & \cellcolor{gray!50} & \cellcolor{gray!50} & \cellcolor{gray!50} & \cellcolor{gray!50} & \cellcolor{gray!50} & \cellcolor{gray!50} & \cellcolor{gray!50} \\
+ MLM (vanilla) & $85.6_{\pm 0.5}$ & $83.9_{\pm 0.2}$ & $81.8_{\pm 0.2}$ & $72.3_{\pm 0.8}$ & $57.0_{\pm 0.3}$ & $34.8_{\pm 2.9}$ & $78.9_{\pm 0.8}$ & $64.1_{\pm 0.2}$ & $48.1_{\pm 2.9}$ \\
+ MLM (whole word) & $86.0_{\pm 0.7}$ & $84.5_{\pm 0.3}$ & $82.7_{\pm 0.4}$ & $72.9_{\pm 0.4}$ & $58.0_{\pm 0.2}$ & $41.6_{\pm 3.3}$ & $79.1_{\pm 0.5}$ & $64.2_{\pm 0.4}$ & $50.1_{\pm 0.9}$ \\
+ MLM (span) & $85.7_{\pm 0.2}$ & $84.1_{\pm 0.1}$ & $82.6_{\pm 0.1}$ & $73.3_{\pm 0.5}$ & $57.9_{\pm 0.8}$ & $39.5_{\pm 2.4}$ & $79.4_{\pm 0.8}$ & $64.3_{\pm 0.4}$* & $54.1_{\pm 1.5}$ \\
+ MLM (entity) & $86.0_{\pm 0.4}$ & $84.3_{\pm 0.3}$ & $82.7_{\pm 0.1}$ & $\mathbf{73.4}_{\pm 0.4}$ & $\mathbf{59.3}_{\pm 1.1}$ & $\mathbf{48.1}_{\pm 4.6}$ & $\mathbf{79.6}_{\pm 0.6}$ & $\mathbf{65.9}_{\pm 1.1}$* & $\mathbf{55.5}_{\pm 2.7}$ \\ \bottomrule
\end{tabular}
}
\caption{EM scores of different continually pretrained models on the original and perturbed test sets. Significant improvements ($p<0.05$) over SpanBERT are marked with *.}
\label{tab:results_defense}
\end{table*}

\section{Related Work}
\label{sec:related}

\paragraph{Robustness of MRC Models.}
The robustness of MRC models are usually evaluated against test-time perturbations and out-of-domain data.
Research on test-time perturbation proposes perturbation methods at different levels as attacks \citep{si2020benchmarking}, such as word replacement with neighbors in the vector space \citep{rychalska2018you, jia2019certified}, question paraphrasing \citep{gan2019improving, ribeiro2018semantically}, sentence distractor injection \citep{jia2017adversarial, zhou2020co}.
Another line of research \citep{fisch2019mrqa,sen2020models} tests a model on data with out-of-domain passage or question distributions, usually from different datasets.
Our work mainly falls into the category of test-time perturbation.
We distinguish from previous work by focusing on the effect of entity renaming, with the motivation that entities can have flexible and diverse names in the real life.

\paragraph{Model Robustness to Entity Substitution.}

It is non-trivial for NLP models to be able to properly handle the large space of named entities.
Previous works use entity substitution to audit or improve model robustness on different tasks like NER \citep{Agarwal2020EntitySwitchedDA, lin2021rockner}, Natural Language Inference \citep{mitra2020enhancing}, Coreference Resolution \citep{subramanian-roth-2019-improving}, and Dialogue State Tracking \citep{cho2021checkdst}.
\citet{shwartz2020you} experiment with name swapping to show that a trained MRC model has bias on some U.S. given names due to the grounding effects that associate names with certain entities.
\citet{ribeiro-etal-2020-beyond} and \citet{balasubramanian2020s} investigate the robustness of models on several tasks with named entity replacement.
However, these works didn't systematically test on MRC datasets with different characteristics to unveil the actual robustness challenge.
\citet{liu2021challenges} study the novel entity generalization ability of open-domain QA models by categorizing the test questions based on whether the named entities have been seen during training.
\citet{longpre2021entity} analyze the memorization behavior of \textit{generative} open-domain QA models using \textit{knowledge conflicts}.
They use entity substitution to create test passages that contain facts contradicting to what the model has learned during training time.
In contrast, we analyze \textit{extractive} MRC model's robustness when encountering new entities, by evaluating on modified test sets without intentionally introduced knowledge conflicts.
The extractive task formulation also makes the model unable to output its memorized knowledge as generative models, leading to different analysis questions and methods.
\section{Conclusion}
\label{sec:conclusion}
In this paper, we systematically study the robustness of MRC models to entity name substitution.
Specifically, we propose a substitution framework along with candidate names of different implications.
We experiment with three pretrained language models on five MRC datasets.
We find that models trained on distantly-supervised datasets are susceptible to entity name substitution, while models trained on human-annotated datasets are relatively robust, with GPE renaming harder than PER and ORG renaming.
The lack of robustness can be further attributed to model's overreliance on entity knowledge and name clues.
We also find that SpanBERT, which is pretrained using span-level objectives, shows better robustness than BERT and RoBERTa.
Leveraging these insights, we study defense approaches based on continual pretraining and demonstrate that entity-based masking policies are beneficial to model's robustness.
Future works include systematically studying the effect of background knowledge in MRC, and developing more effective methods to improve the robustness of MRC models.
\section*{Acknowledgments}
This research is supported in part by the Office of the Director of National Intelligence (ODNI), Intelligence Advanced Research Projects Activity (IARPA), via Contract No. 2019-19051600007, the DARPA MCS program under Contract No. N660011924033, the Defense Advanced Research Projects Agency with award W911NF-19-20271, NSF IIS 2048211, NSF SMA 1829268, and gift awards from Google, Amazon, JP Morgan and Sony. We would like to thank all the collaborators in USC INK research lab for their constructive feedback on the work.
We would also like to thank the anonymous reviewers for their valuable comments.

\bibliography{custom}
\bibliographystyle{acl_natbib}

\clearpage
\appendix
\section{Details for MRC Model Training}
\label{app:training}

We train all MRC models using mixed precision, with batch size of 16 sequences for 4 epochs.
The maximum sequence length is set to 256 tokens.
We use the AdamW optimizer \citep{loshchilov2017decoupled} with an initial learning rate of 2e-5 that is linearly decayed to 0 during finetuning.

\section{Perturbable Span Identification}
\label{app:perturbable}
For PER, we only consider names with one or two words.
A one-word name is considered as a first name, while a two-word name is considered as a full name, with the first word being the first name and the second word being the last name.
We infer the gender of the detected name to be male, female, or neutral with \texttt{gender-guesser}\footnote{\scriptsize\url{https://pypi.org/project/gender-guesser/}}.

For GPE, we detect its contained country names, state names, and city names by string matching with the \textit{Countries States Cities Database}\footnote{\scriptsize\url{https://countrystatecity.in/}}.

For ORG, besides mentions of GPE names, we include two additional types of perturbable words identified using Penn Treebank (PTB) \citep{marcus1993building}.
Words that are annotated as NNP(S) for more than 90\% of the time in PTB are considered as proper nouns (dented as <NNP>), which are usually specialized for naming an entity.
Words outside PTB are considered as rare words (denoted as <Rare>), which are likely to be invented by people to name an entity.
These two kinds of words are weakly related to the characteristics of the entity and thus can be flexible.

\section{Evaluation of Perturbation Quality}
\label{app:eval_perturbation}
The accuracy of step 2 is evaluated based on whether the perturbable spans and their corresponding span types are all correct for an instance, which also implies the quality of step 1 (``Answer Entity Recognition'') as different entity types have different applicable span types.

The accuracy of step 4 is evaluated based on whether the string mapping function successfully locates all mentions of the perturbable spans in the passage to perform string mapping.

The quality of step 3 can be inferred from the accuracy of step 2 for InDistName perturbation.
For DBName, we assume the database is of acceptable quality in the sense that all names it provides belongs to the correct span type, which is guaranteed by the source of the data---PTB is annotated by human experts, U.S. names come from official statistics, and GPE names are actively maintained by its creator and the community for more than 3 years.
RandStr is proposed to simulate the extreme case, and we therefore do not evaluate its quality.

\section{Details for Continual Pretraining}
\label{app:continual}

\textbf{MLM (vanilla)} refers to the masking strategy used by BERT \citep{devlin2018bert}, where the masked tokens are randomly sampled.
\textbf{MLM (whole word)} always masks all tokens corresponding to a word at once.
\textbf{MLM (span)} uses the masking strategy proposed by \citet{joshi2020spanbert}, which masks random spans rather than individual whole words or tokens. 
\textbf{MLM (entity)} masks a random entity for 50\% of the time, and uses MLM (span) for the other 50\% of the time.
The idea is inspired by salient span masking proposed in \citet{guu2020realm}.
All strategies mask 15\% of the training tokens in total.

To eliminate domain shift during continual pretraining as a possible explanation for any improvements, we keep the corpus for continual pretraining consistent with the pretraining corpus used by SpanBERT, which is the concatenation of BookCorpus and English Wikipedia.
We train using mixed precision, with effective batch size of 2,048 sequences for 8,000 steps, with 256 tokens per sequence.
We use the AdamW optimizer with a constant learning rate of 1e-4.

\end{document}